\documentclass[a4paper,fleqn,svgnames,dvipsnames,table]{cas-dc}

\usepackage[numbers, sort&compress]{natbib}
\usepackage{caption}
\usepackage{bbding} 
\usepackage{amsmath}
\usepackage{hyperref}

\usepackage{graphicx}
\usepackage{bbding}
\usepackage{amsmath}
\usepackage{amsfonts}
\usepackage{enumerate}
\usepackage{booktabs}
\usepackage{algorithm}
\usepackage{algorithmic}
\usepackage{color}
\usepackage{colortbl}
\usepackage{bbm}
\usepackage{multirow}
\usepackage{array}
\usepackage{pifont}
\usepackage{xcolor}

\definecolor{greenc}{RGB}{0,176,80}
\newcommand{\greenc}[1]{{\color{greenc}#1}}

\newcommand{\etal}{\textit{et al.}}

\makeatletter
\newcommand{\removelatexerror}{\let\@latex@error\@gobble}
\makeatother

\def\tsc#1{\csdef{#1}{\textsc{\lowercase{#1}}\xspace}}
\tsc{WGM}
\tsc{QE}

\begin{document}
\let\WriteBookmarks\relax
\def\floatpagepagefraction{1}
\def\textpagefraction{.001}

\shorttitle{}    
\shortauthors{Xu et al.}  

\title[mode = title]{MambaMoE: Mixture-of-Spectral-Spatial-Experts State Space Model for Hyperspectral Image Classification}



\author[1]{Yichu Xu}[style=chinese]
\ead{xuyichu@whu.edu.cn}

\address[1]{School of Computer Science, Wuhan University, Wuhan, 430072, PR China}
\address[2]{School of Urban Design, Wuhan University, Wuhan, 430072, PR China}
\address[3]{Aerospace Information Research Institute, Henan Academy of Sciences, Zhengzhou 450046, PR China}

\author[1]{Di Wang}[style=chinese]
\ead{d\_wang@whu.edu.cn}


\author[2]{Hongzan Jiao}[style=chinese]
\ead{jiaohongzan@whu.edu.cn}

\author[1]{Lefei Zhang}[style=chinese]
\cormark[1]
\credit{Supervision.}
\ead{zhanglefei@whu.edu.cn}

\author[3]{Liangpei Zhang}[style=chinese]
\ead{zlp62@whu.edu.cn}

\cortext[1]{Corresponding author.}

\begin{abstract}
Mamba-based models have recently demonstrated significant potential in hyperspectral image (HSI) classification, primarily due to their ability to perform contextual modeling with linear computational complexity.  
However, existing Mamba-based approaches often overlook the directional modeling heterogeneity across different land-cover types, leading to limited classification
performance. To address these limitations, we propose MambaMoE, a novel spectral-spatial Mixture-of-Experts (MoE) framework, which represents the first MoE-based approach in the HSI classification domain. Specifically, we design a Mixture of Mamba Expert Block (MoMEB) that performs adaptive spectral-spatial feature modeling via a sparse expert activation mechanism. Additionally, we introduce an uncertainty-guided corrective learning (UGCL) strategy that encourages the model to focus on complex regions prone to prediction ambiguity. This strategy dynamically samples supervision signals from regions with high predictive uncertainty, guiding the model to adaptively refine feature representations and thereby enhancing its focus on challenging areas. Extensive experiments conducted on multiple public HSI benchmark datasets show that MambaMoE achieves state-of-the-art performance in both classification accuracy and computational efficiency compared to existing advanced methods, particularly Mamba-based ones. The code will be available online at https://github.com/YichuXu/MambaMoE.
\end{abstract}



\maketitle
\begin{sloppypar}

\section{Introduction}\label{introduction}
{Hyperspectral images (HSIs) contain hundreds of contiguous spectral bands, enabling the capture of rich spectral-spatial information that supports accurate analysis for Earth observation tasks \cite{SONG2025DL,fu2024hyperdehazing,Wu24HSIC-IF,Tu24IF,hypersigma}. HSI classification, which involves assigning a semantic label to each pixel in the image, has emerged as a fundamental task in the remote sensing community \cite{kang24IF,LU2023CAL,fu2023three}.}

{With the rise of deep learning, HSI classification has transitioned from handcrafted feature engineering methods, such as tensor singular spectrum analysis \cite{SVM,RForest,fu2023tensor}, to fully automated neural network-based approaches.} Early methods, such as CNNs \cite{hu2015deep,2D-CNN}, demonstrated strong local feature extraction capabilities but were limited in capturing global context due to their constrained receptive fields. To address this, Transformer-based models \cite{SpectralFormer,LI2024CasFormer,DSFormer} introduced long-range dependency modeling. However, their reliance on the self-attention mechanism \cite{selfattention,Xu2024s3anet} incurs quadratic computational complexity, making them less efficient for modeling spatial and spectral contexts simultaneously.

Recently, Mamba-based architectures have garnered increasing attention in HSI classification due to their context modeling capability while maintaining linear complexity \cite{gu2021efficiently,gu2023mamba}. However, existing approaches exhibit several limitations. Firstly, most existing methods adopt patch-level designs \cite{s2mamba, mambaLG}, where an image patch is fed into the network to predict the category of the central pixel. This approach restricts the receptive field and fails to capture the holistic spatial context of remote sensing scenes. In addition, although multi-directional spectral-spatial scanning \cite{hypermamba} is commonly employed to enhance feature diversity, existing methods typically treat all directions equally, overlooking the directional variations inherent in heterogeneous land-cover types (see Section \ref{vis}). 
Furthermore, the inherently sequential modeling \cite{xiao2024fmsr} paradigm of Mamba disrupts the original spatial structure of hyperspectral data, resulting in a loss of spatial coherence. This degradation limits the model’s ability to preserve and exploit critical spatial information, ultimately hindering the extraction of fine-granularity and context-aware features that are essential for accurate classification.

To address these limitations, inspired by the Mixture-of-Experts (MoE) paradigm \cite{zhang2025uniuir,RingMoE}, we propose MambaMoE—a spectral-spatial mixture-of-mamba-expert framework for HSI classification. This framework is composed of specialized network blocks based on Mamba structures, which leverage state space models oriented along distinct spatial and spectral directions. Within these blocks, multiple Mamba modules serve as diverse experts, collectively forming the proposed Mixture of Mamba Expert Block (MoMEB). Given that remote sensing images are captured from a bird's-eye view—where objects can appear in arbitrary orientations—and hyperspectral data channels follow a fixed wavelength order, i.e., exhibiting directional spectral dependencies, MoMEB is designed to reflect these spatial-spectral characteristics. Specifically, it treats each spatial-directional Mamba branch as a routed expert and employs two spectral-directional branches as shared experts. Furthermore, a lightweight router network is integrated within MoMEB to dynamically activate a subset of experts for each input, enabling the model to adaptively extract informative spectral-spatial directional features. This design significantly enhances classification performance while maintaining computational efficiency.

Different from existing patch-level Mamba classification networks \cite{huang-SSMamba,hypermamba}, MambaMoE is an end-to-end image-level segmentation network built upon a classical encoder-decoder architecture. As described above, MoMEBs are employed in the encoder to extract representative spectral-spatial features. However, due to the limited directional diversity in Mamba's sequential modeling, these features may not fully capture the varied object orientations present in complex HSI scenes, potentially leading to classification ambiguity. To mitigate this, we introduce uncertainty-guided corrective learning (UGCL) in the decoder. Specifically, UGCL encourages the model to focus on challenging regions by dynamically sampling supervisory labels from areas of high prediction uncertainty, guiding MoMEB outputs to adaptively refine their representations. This strategy enables the network to better handle difficult cases and enhances its generalization capability. Thanks to this design, MambaMoE achieves state-of-the-art performance across multiple public HSI benchmarks, surpassing existing advanced methods, including those based on Mamba architectures.

The main contributions of this work can be summarized as follows:

\begin{itemize}
\item[1)] We propose MambaMoE, a novel Mamba-based spectral-spatial mixture-of-experts framework for HSI classification. To the best of our knowledge, this is the first MoE-based deep network introduced in the HSI classification domain, enabling adaptive extraction of spectral-spatial joint features tailored to the diverse characteristics of land covers.

\item[2)] We design the Mixture of Mamba Expert Block (MoMEB), which integrates spatially routed experts and spectrally shared experts based on Mamba. Leveraging sparse expert activation, MoMEB facilitates dynamic learning of directional spectral-spatial features.

\item[3)] We introduce an uncertainty-guided corrective learning (UGCL) strategy to guide the model’s focus toward challenging regions. This approach mitigates prediction confusion arising from the directional modeling limitations of Mamba by adaptively refining feature representations in uncertain areas.

\item[4)] Experiments on multiple HSI classification benchmarks demonstrate that our proposed method consistently outperforms existing state-of-the-art approaches in both accuracy and computational efficiency, thanks to the synergy of our architecture and training strategy.

\end{itemize}


\par The remaining section of this paper is organized as follows. Section \ref{Section2} reviews the related work. Section \ref{Section3} provides a comprehensive description of the proposed MambaMoE model. Section~\ref{Section4} provides a comprehensive presentation and analysis of the experimental results. Finally, Section \ref{Section5} concludes the paper.

\section{Related Work} \label{Section2}
\subsection{Mamba-based HSI Classification}
Recently, Mamba has attracted increasing attention in the field of HSI classification due to its ability to achieve context modeling with linear computational complexity. Yao~\etal~\cite{yao2024spectralmamba} proposed SpectralMamba, which simplifies sequence learning in the state space and enhances spectral correction within the spectral-spatial domain. SS-Mamba~\cite{huang-SSMamba} further extends this idea by introducing multiple spectral-spatial Mamba modules. In addition, Li~\etal~\cite{MambaHSI} proposed MambaHSI, which models long-range dependencies across the entire image and adaptively fuses spatial and spectral information. HyperMamba~\cite{hypermamba} introduces a spectral-spatial adaptive architecture that jointly performs spatial neighborhood scanning and spectral feature enhancement. He~\etal~\cite{3DSSMamba} explored the application of Mamba from a 3D perspective, focusing on 3D cubes. MambaLG~\cite{mambaLG} employs a dual-branch architecture that integrates local-global spatial modeling and dynamic short- and long-range spectral feature perception, fully leveraging Mamba’s sequence modeling capabilities. Wang~\etal~\cite{s2mamba} proposed S2Mamba, a spectral-spatial scanning-based model that employs a learnable mixture gate to effectively fuse spatial and spectral representations. Building upon this line of work, our method also adopts image-level input, as used in MambaHSI~\cite{MambaHSI}, but further enhances modeling capacity by integrating a sparsely activated MoE in the encoder, enabling dynamic and adaptive spectral-spatial representation. Additionally, we design the UGCL strategy to dynamically guide the model to learn from labels sampled in complex regions, effectively mitigating classification ambiguities in challenging areas.

\subsection{Mixture-of-Experts} 
Mixture-of-Experts (MoE) have emerged as a promising paradigm for enhancing model efficiency and scalability by enabling dynamic, data-dependent routing through specialized sub-networks (experts). MoE has been widely adopted in the NLP community to scale large language models, as exemplified by MoA~\cite{MoA}, DeepseekMoE~\cite{deepseekmoe} and others \cite{zhou2022mixture}. Beyond NLP, MoE has also gained traction in the computer vision domain. For instance, Mod-Squad~\cite{Mod-Squad} utilizes MoE for efficient multi-task learning, while RingMoE~\cite{RingMoE} adopts a sparse MoE design to build a unified multi-modal foundation model for remote sensing. Despite its success in both NLP and vision tasks, its application to HSI classification remains largely unexplored. In this work, we introduce the MoE into HSI classification for the first time, enabling adaptive spectral-spatial joint feature learning through the proposed spatially routed experts and spectrally shared experts.
\section{Methodology}\label{Section3}

\subsection{Overview}
\begin{figure*}[t]
    \centering
     \includegraphics[width=\linewidth]{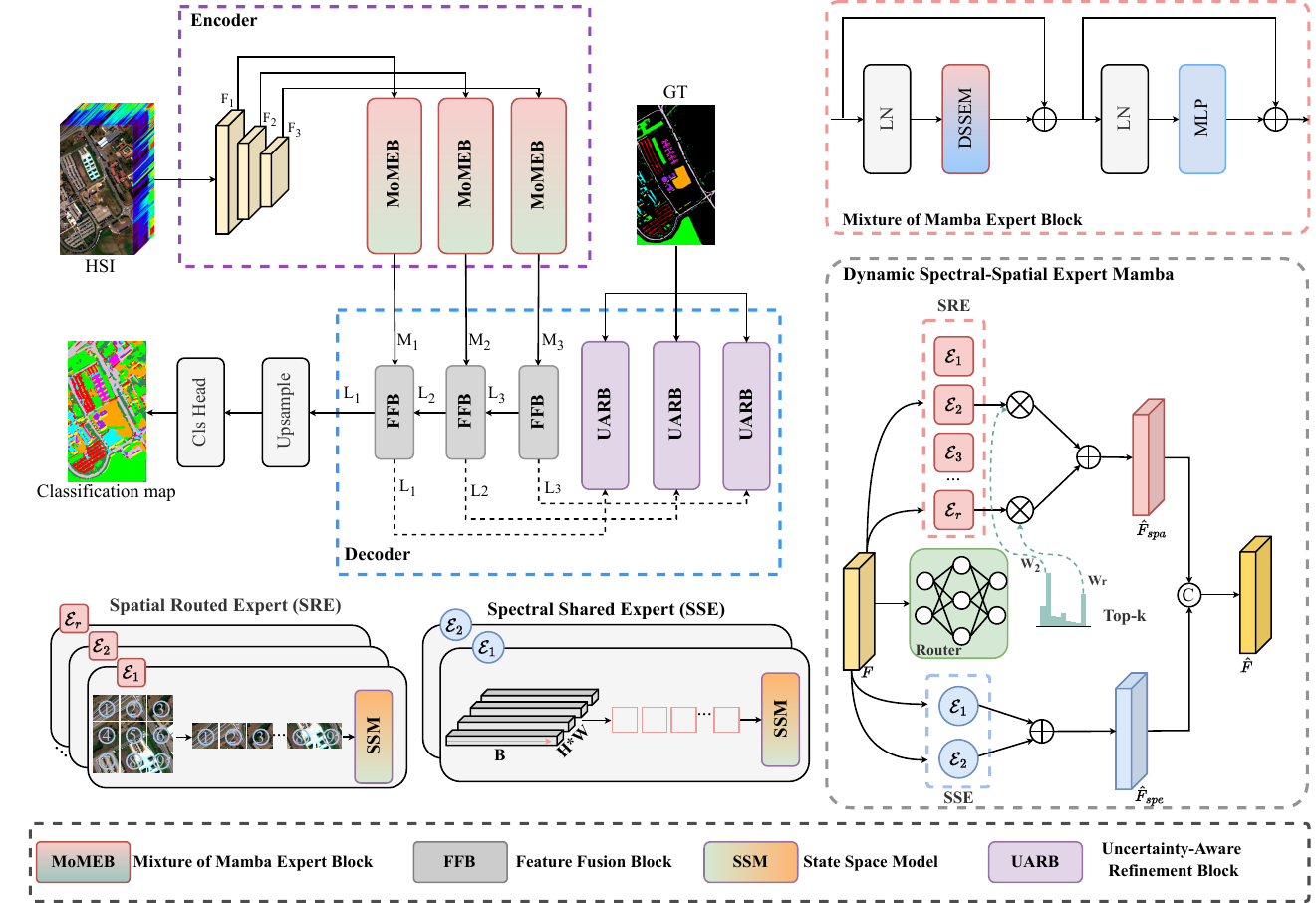}
     \caption{An illustration of the proposed MambaMoE. The core component, MoMEB incorporates DSSEM, which leverages SRE and SSE to dynamically capture spectral-spatial features. The Decoder is composed of FFBs and UARBs. After feature extraction and refinement, we adopt the fully connected layer as the classification head to generate the final predictions.}
     \label{fig:framework}
\end{figure*}

As illustrated in Fig. \ref{fig:framework}, the entire HSI $\mathbf{X} \in \mathbb{R}^{B \times H \times W}$ is processed by the encoder-decoder segmentation network, MambaMoE. Here, $H$ and $W$ represent the height and width, respectively, and $B$ denotes the number of spectral bands in the HSI. The encoder comprises a feature extraction module followed by multiple MoMEB blocks. Initially, hierarchical features $\mathbf{F}_i \in \mathbb{R}^{C \times H_{i} \times W_{i}} i=1,2,3$ are extracted via the feature extraction module, formulated as: 
\begin{equation}
\begin{split}
&\mathbf{F}_{1}=Avg(ReLU(Conv(\mathbf{X})))\\
&\mathbf{F}_{i}=Avg(ReLU(Conv(\mathbf{F}_{i-1}))), i=2,3. 
\end{split}
\end{equation}
Here, $Conv(\cdot)$, $GN(\cdot)$, $ReLU(\cdot)$, and $Avg(\cdot)$ denote a $3 \times 3$ convolution operation, a ReLU activation function, and an average pooling layer, respectively. Subsequently, these features are fed into the proposed MoMEB modules to adaptively capture spectral-spatial contexts, defined as: $\mathbf{M}_i=\text{MoMEB}(\mathbf{F}_i), \mathbf{M}_i \in \mathbb{R}^{C \times H_{i} \times W_{i}}, i=1,2,3$. In the decoder, a group of Feature Fusion Blocks (FFBs) are applied to integrate inter-scale features from different stages, which can be formulated as: 
\begin{equation}
\mathbf{L}_i = 
\begin{cases}
\text{FFB}(\mathbf{M}_i, \mathbf{L}_{i+1}), & i = 1, 2 \\
\text{FFB}(\mathbf{M}_i), & i = 3
\end{cases}
\label{eqn:H}
\end{equation}
Here, $\mathbf{L}_i \in \mathbb{R}^{C \times H_{i} \times W_{i}}, i=1,2,3$. Then, we apply the UGCL, in which the feature $\mathbf{L}_i$ and the ground-truth labels are passed through the Uncertainty-Aware Refinement Block (UARB) to generate sampled supervision labels for complex regions. This process is formally expressed as: $\mathbf{Q}_i = \text{UARB}(\mathbf{L}_i) \in \mathbb{R}^{H \times W}$. Notably, UARB is applied only during training to enhance learning on uncertain regions, and is omitted during inference.

\noindent\textbf{Feature Fusion Block} In the Feature Fusion Block (FFB), the feature map $\mathbf{M}_3$, obtained from the MoMEB module, is first processed through a residual structure to produce $\mathbf{L}_3$: $\mathbf{L}_3 = \text{Res}(\mathbf{M}_3)$, where the residual function is defined as: 
\begin{equation}
    \text{Res}(x) = x + \text{Conv2}_{3\times 3}(\text{ReLU}(\text{Conv1}_{3\times 3}(\text{ReLU}(x))))
\end{equation}
with two sequential $3 \times 3$ convolution layers. In the following stages, multi-scale feature fusion is achieved by combining the upsampled output from the previous FFB block, $\mathbf{L}_{i+1}$, with the current feature map $\mathbf{M}_i$. This is formulated as: $\mathbf{L}_{i} = \mathbf{M}_i+\text{Res}(\text{Up}(\mathbf{L}_{i+1}))$, where $\text{Up}(\cdot)$ denotes bilinear interpolation-based upsampling.

\subsection{Mixture of Mamba Expert Block}

{The MoMEB module comprises two key sub-components: the Spatial Routed Expert (SRE) and the Spectral Shared Expert (SSE).} The SRE leverages spatial Mamba structures with distinct scanning directions to dynamically capture essential spatial context. In contrast, the SSE is designed using spectral Mamba structures with both forward and backward scanning directions, enabling the extraction of shared spectral representations from the HSI. 

Given an input feature map $\mathbf{F}$ (the stage index is omitted in later texts for convenience), it is first normalized by a LayerNorm (LN) layer, obtaining a new feature $\overline{\mathbf{F}}$. Next, this feature is processed by the Dynamic Spectral-Spatial Expert Mamba (DSSEM) module. Specifically, the input feature is split along the channel dimension into two equal parts, forming a spatial view $\mathbf{F}_{spa} \in \mathbb{R}^{(C/2) \times h \times w}$ and a spectral view $\mathbf{F}_{spe} \in \mathbb{R}^{(C/2) \times h \times w}$, where $h$ and $w$ are height and width of the feature in corresponding stages. These two views are separately fed into the SRE and SSE modules to extract spatial and spectral contextual features, respectively. The resulting outputs, $\text{SRE}(\mathbf{F}_{spa})$ and $\text{SSE}(\mathbf{F}_{spe})$, are concatenated and then passed through a $1 \times 1$ convolution to integrate the information, producing a dynamically modulated spectral-spatial fusion feature $\hat{\mathbf{F}}$. The overall procedure of DSSEM is
\begin{equation}
\begin{split}
    &\mathbf{F}_{spa}, \mathbf{F}_{spe} = \text{ChannelSplit}(\overline{\mathbf{F}})\\
    &\hat{\mathbf{F}} = Conv_{1 \times 1}(\text{Concat}(\text{SRE}(\mathbf{F}_{spa}), \text{SSE}(\mathbf{F}_{spe})))
\end{split}
\end{equation}

This fused feature $\hat{\mathbf{F}}$ is then combined with the original input $\mathbf{F}$ via a residual connection. Subsequently, the output is further refined through a Layer Normalization (LN) layer followed by a Multi-Layer Perceptron (MLP) layer, and again enhanced by residual addition. Overall, this process within MoMEB resembles the structure of a standard Transformer block, where the attention mechanism is replaced by the DSSEM module, with additional split-and-merge operations on the input and output features of DSSEM.

\noindent\textbf{Spatial Routed Expert.} The SRE module consists of multiple Mamba experts, each configured with a distinct spatial scanning direction. Each expert employs a dedicated state space model (SSM) to capture directional structural information from the input. In this work, we specifically adopt four canonical scanning directions: top-left to bottom-right, bottom-right to top-left, top-right to bottom-left, and bottom-left to top-right, allowing the model to capture spatial dependencies from different perspectives. Specifically, for each expert, the spatial feature $\mathbf{F}_{spa}$ is first flattened into a one-dimensional sequence according to a predefined scanning path. The resulting sequence for the $j$-th expert is denoted as:
\begin{equation}
F_{spa}^j = \left\{ [f_j^1, f_j^2, \cdots, f_j^{h \times w}] \;\middle|\; f_j \in \mathbb{R}^{1 \times (C/2)},\; j \in \{1, 2, 3, 4\} \right\}
\end{equation}
which is then processed by a SSM as follows:
\begin{equation}
\begin{aligned}
&\mathbf{h}_j^t = \overline{\mathbf{A}}^{spa} \mathbf{h}_j^{t-1} + \overline{\mathbf{B}}^{spa} f_j^t, \\
&\mathbf{y}_j^t = \mathbf{C}^{spa} \mathbf{h}_j^t + f_j^t,
\end{aligned}
\end{equation}

where $\mathbf{y}_j^t \in \mathbb{R}^{1 \times (C/2)}, t = 1,\cdots, h \cdot w $ is the output of each pixel location, $\overline{\mathbf{A}}^{spa}\in \mathbb{R}^{D \times D}$ is the state transition matrix, $\overline{\mathbf{B}}_{i,k}^{spa}\in \mathbb{R}^{D \times (C/2)}$ and $\mathbf{C}^{spa} \in \mathbb{R}^{(C/2) \times D}$ are projection matrices, all matrics are trainable. $D$ denotes the dimensionality of the latent state space. The index $j$ identifies the specific expert corresponding to one of the four canonical scanning directions. It can be seen that, each expert is able to capture unique spatial structural patterns according to its designated scanning direction. The whole output of expert $\mathcal{E}_j, j=1,2,3,4$ is $\left[\mathbf{y}_j^1, \mathbf{y}_j^2, \dots, \mathbf{y}_j^{h \times w} \right] \in \mathbb{R}^{(C/2)\times h \times w}$.

To adaptively capture directional context for heterogeneous objects in the HSI scene, we introduce a router network for dynamic expert routing, denoted as $\mathcal{G}$. This network comprises two MLP layers and is designed to select the most suitable experts based on the input features by generating a set of weighting coefficients:
\begin{equation}
\mathbf{w} = \text{softmax}(\mathcal{G}(\mathbf{F}_{spa}))
\end{equation}
where the softmax function ensures that the weights assigned to each expert sum to one. Using these weights, the final output of the SRE module, denoted as $\mathbf{\hat{F}}_{spa}$, is computed as a weighted combination of the expert outputs:
\begin{equation}
\mathbf{\hat{F}}_{spa} = \sum_{j=1}^{4} \mathbf{w}_j \cdot \mathcal{E}_j(\mathbf{F}_{spa}).
\end{equation}
This dynamic routing mechanism enables the model to emphasize the most relevant directional features for each objects, enhancing its ability to model complex spatial structures in HSIs.

During training, all experts participate in learning to ensure diverse directional representations are captured. However, to improve computational efficiency during inference, we leverage the sparsity of the routing function $\mathcal{G}(\cdot)$. Specifically, for each input, only the top-$k$ experts with the highest routing weights are selected, and their outputs are used to compute the final result. This selective inference strategy balances model performance with efficiency by focusing computation on the most relevant experts.

\noindent\textbf{Spectral Shared Expert.} { The SSE module performs essential spectral context modeling and functions as a shared spectral feature extractor, enabling the model to capture consistent spectral patterns across different spatial locations. In the SSE module, forward and backward spectral sequences are constructed along the spectral dimension and processed using SSMs to extract bi-directional spectral features.} As illustrated in Fig.~\ref{fig:framework}, the input feature $\mathbf{F}_{spe}$ is first flattened along the spatial dimensions to form spectral sequences:
\begin{equation}
    F^{l}_{spe} = \left\{ [\Tilde{f}_l^0, \Tilde{f}_l^1, \cdots, \Tilde{f}_l^{(C/2)}] \;\middle|\; \Tilde{f}_l \in \mathbb{R}^{1 \times (h\cdot w)},\; l \in \{1, 2\} \right\}.
\end{equation}
where $l=1$ and $l=2$ correspond to the forward and backward scanning directions, respectively. Each sequence is then passed through an SSM to obtain the corresponding spectral outputs, formulated as:
\begin{equation}
\begin{aligned}
&\mathbf{\Tilde{h}}_l^t = \overline{\mathbf{A}}^{spe} \mathbf{\Tilde{h}}_l^{t-1} + \overline{\mathbf{B}}^{spe} \Tilde{f}_l^t, \\
&\mathbf{\Tilde{y}}_l^t = \mathbf{C}^{spe} \mathbf{\Tilde{h}}_l^t + \Tilde{f}_l^t,
\end{aligned}
\end{equation}
Here, $\mathbf{\Tilde{y}}_l^t \in \mathbb{R}^{1 \times (h \cdot w)}$ represents the $t$-th output in the $l$-th directional sequence. The matrices $\overline{\mathbf{A}}^{spe} \in \mathbb{R}^{D \times D}$, $\overline{\mathbf{B}}^{spe} \in \mathbb{R}^{D \times (h \cdot w)}$, and $\mathbf{C}^{spe} \in \mathbb{R}^{(h \cdot w) \times (h \cdot w)}$ are trainable parameters, and $D$ denotes the dimensionality of the latent state, which is the same as SRE. 

Finally, the outputs from the forward and backward directions are additively fused to form the final bi-directional spectral representation $\mathbf{\hat{F}}_{spe} \in \mathbb{R}^{(C/2) \times h \times w}$. This fusion incorporates contextual information from adjacent spectral bands in both directions, thereby enhancing the model’s spectral modeling capability.

\begin{figure}[t]
    \centering
     \includegraphics[width=\linewidth]{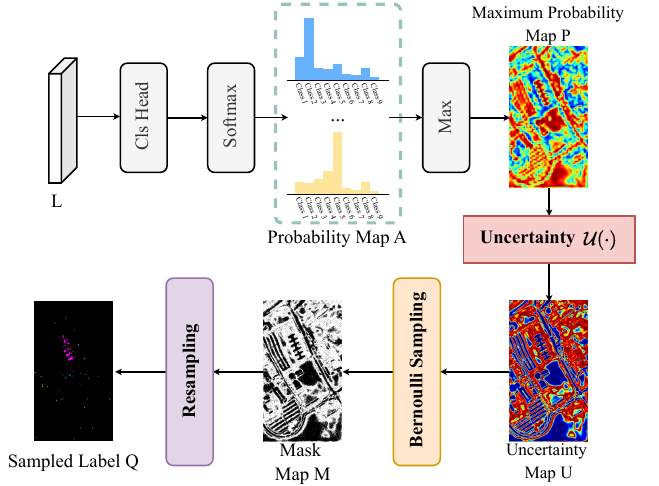}
     \caption{The overview of the UARB.}
     \label{UARB}
\end{figure}

\subsection{Uncertainty-Guided Corrective Learning}

In addition to the MoMEBs and inspired by \cite{li2025overcoming}, we further enhance predictive performance in challenging regions by encouraging the model to learn from dynamically identified difficult areas, as determined by uncertainty estimation. To this end, we propose the UARB, as illustrated in Fig.~\ref{UARB}. The UARB processes intermediate features $\mathbf{L} \in \mathbb{R}^{C \times h \times w}$ from various decoder stages. 

Specifically, $\mathbf{L}$ is passed through a classification head to produce raw prediction scores $\mathbf{S} \in \mathbb{R}^{{n_{class}} \times h \times w}$, where $n_{class}$ denotes the number of semantic categories. These scores are normalized via the softmax function to generate a class-wise probability matrix:

\begin{equation}
\mathbf{A}_{k}= \frac{\exp(\mathbf{S}_{k})}{\sum\limits_{j=1}^{n_{class}} \exp(\mathbf{S}_{j})},
\end{equation}
where $\mathbf{S}_k \in \mathbb{R}^{h \times w}$ represents the score map for the $k$-th class, and $\mathbf{A}_k$ is its corresponding probability map.

Intuitively, the uncertainty of a pixel is inversely correlated with the model's confidence in its prediction. To quantify this, we compute the maximum class probability for each pixel by taking the maximum value across the class dimension:
\begin{equation}
\mathbf{P}_{r,c} = \max\limits_{k \in \{1, \dots, n_{class}\}} \mathbf{A}_{k,r,c}, \quad \mathbf{P}_{r,c} \in [0, 1].
\end{equation}
where $r \in [0, h-1]$ and $c \in [0, w-1]$ denote spatial coordinates.

To derive a pixel-wise uncertainty map $\mathbf{U}$, we apply an uncertainty estimation function $\mathcal{U}(\cdot)$ on $\mathbf{P}$:
\begin{equation}
\mathbf{U}_{r,c} = -\log(\mathbf{P}_{r,c} + \epsilon) \cdot \mathbf{P}_{r,c},
\end{equation}
where $\epsilon = 10^{-6}$ is a small constant to prevent numerical instability. The resulting uncertainty map $\mathbf{U} \in [0, 1]$ reflects the confidence distribution of the model across spatial positions. Notably, regions with higher uncertainty, typically associated with complex or ambiguous features, are emphasized for refinement.

To guide the model's attention toward uncertain regions, we aim to obtain and leverage the labels of these areas during training. Intuitively, diverse and informative labels from these regions can enhance the model's robustness. To this end, we employ probabilistic sampling to select pixel locations for supervision, ensuring that regions with higher uncertainty are more likely to be sampled.

Specifically, we adopt Bernoulli sampling at each pixel location of the uncertainty map $\mathbf{U}$. The sampling process is defined as:
\begin{equation}
    M_{r,c} = 
\begin{cases} 
1 & \text{if } r_{r,c}  < \mathbf{U}_{r,c} \\
0 & \text{if } r_{r,c}  \geq \mathbf{U}_{r,c},
\end{cases}
\end{equation}
where $M \in {0,1}^{h \times w}$ is a binary mask map indicating which spatial locations are selected for label supervision, and $r_{r,c}$ denotes a random value drawn from a uniform distribution $U(0, 1)$ at pixel $(r, c)$.

In practice, this sampling is conducted dynamically at each training iteration, allowing the model to adaptively focus on different uncertain regions as learning progresses. The final sampled label map $\mathbf{Q}_i, i=1,2,3$ is obtained by applying an element-wise product between the binary mask $M$ and the ground-truth label map $\mathbf{Y}_{trn}$. Notably, for each stage, $\mathbf{L}_{i}$ is accordingly upsampled to match the spatial size of $\mathbf{Q}_i$.

\subsection{ Training Loss}

In this study, since image-level HSI classification is formulated as a multi-class segmentation task, we adopt the cross-entropy loss function as the objective criterion. The total training loss comprises two components: (1) the losses computed at each intermediate stage using the sampled supervision labels, and (2) the loss from the final prediction, which is computed against the original ground-truth labels. All loss terms are weighted equally, with a coefficient of 1. Therefore the overall loss function of this work $\mathcal{L}$ is defined as follows:

\begin{equation}
 \mathcal{L} = \sum_{i=1}^{3} \mathcal{L}_{ce}(\mathbf{Q}_{i}, \mathbf{A}_{i}) + \mathcal{L}_{ce}(\mathbf{GT}, \mathbf{A}_{3}),
\label{eqn:cls_loss}
\end{equation}
where $\mathcal{L}_{ce}$ denotes the cross-entropy loss function, $\mathbf{Q}_{i}$, $\mathbf{A}_{i}$, and GT correspond to the sampled labels, the predicted probability, and the training labels, respectively.

\section{Experiment}\label{Section4}
\subsection{Dataset}

We employ three benchmark hyperspectral datasets to evaluate the effectiveness of the proposed method. A detailed description is provided below. 

\begin{enumerate}[(1)]
\item \textbf{Pavia University Dataset.} This dataset is captured by the Reflective Optics System Imaging Spectrometer (ROSIS) in 2001, this dataset covers the University of Pavia in northern Italy. It includes 103 spectral bands and a spatial resolution of 610 × 340 pixels, encompassing nine land-cover classes.

\item \textbf{Houston Dataset.} This dataset is acquired in 2012 over the University of Houston and its surrounding urban areas, this dataset consists of 144 spectral bands within the 400–1000 nm wavelength range. It has a spatial resolution of 2.5 meters, an image size of 349 × 1905 pixels, and includes 15 semantic categories \cite{houston2013}.

\item \textbf{Whu-HanChuan Dataset.} This dataset is introduced by Zhong et al. \cite{zhong2020whu}, this dataset was obtained via UAV-borne imaging, offering rich spatial detail with centimeter-level resolution. It features 274 spectral bands within the 400–1000 nm range and a spatial size of 1217 × 303 pixels.
\end{enumerate}

\subsection{Experimental Settings}The experiments in this study were performed using the PyTorch framework, with an NVIDIA GeForce RTX 3090 GPU for computation. For the first two datasets, \textbf{only 15 samples per class} were randomly chosen for the training set, while \textbf{30 samples per class} were selected for the Whu-HanChuan dataset. The remaining labeled samples were used exclusively as the test set. The training process employed the Adam optimizer with a learning rate of $5 \times 10^{-4}$ and was conducted for 200 epochs across all datasets. To quantitatively evaluate classification performance, three widely recognized metrics were employed: overall accuracy (OA), average accuracy (AA), and the kappa coefficient ($\kappa$). Each experiment was repeated 10 times with different random seeds, and the results are reported as the mean and standard deviation.

\begin{table}[t]
\caption{Performance contribution of different module on three benchmark datasets (reported in OA (\%). Best results are highlighted in \textbf{Bold}). }
\centering
\resizebox{\linewidth}{!}{
\begin{tabular}{cc|ccc} %
\toprule
MoMEB & UARB   &    Pavia University & Houston & Whu-HanChuan \\
\midrule
\XSolidBrush & \XSolidBrush  & 91.53 & 89.47 & 89.17 \\
\Checkmark & \XSolidBrush    & 93.86 \greenc{($\uparrow$ 2.33)} & 90.49 \greenc{($\uparrow$ 1.02)} & 90.86 \greenc{($\uparrow$ 1.69)} \\
\XSolidBrush  & \Checkmark     & 92.78 \greenc{($\uparrow$ 1.25)} & 90.25 \greenc{($\uparrow$ 0.78)} & 89.55 \greenc{($\uparrow$ 0.38)}\\
\rowcolor{gray!30} \Checkmark & \Checkmark  &\textbf{95.20} \greenc{($\uparrow$ 3.67)}&\textbf{91.18} \greenc{($\uparrow$ 1.71)}&\textbf{92.67} \greenc{($\uparrow$ 3.50)}\\
\bottomrule
\end{tabular}
}
\label{table:ablation}
\end{table}

\subsection{Ablation Study}

\noindent\textbf{(1) Network Blocks.} To validate the effectiveness of the proposed MoMEB and UGCL, we conduct ablation studies on three HSI datasets. The baseline model includes only multi-scale feature extraction and stage-wise fusion, without the incorporation of either MoMEB or UARB. We evaluate the impact of integrating MoMEB and UARB individually and jointly. As shown in Table~\ref{table:ablation}, both modules contribute significantly to performance improvement. On the Pavia University dataset, incorporating MoMEB improves the baseline OA by 2.33\%, while UARB yields a 1.25\% increase. When both modules are applied together, the model achieves the highest OA of 95.20\%, representing an overall improvement of 3.67\% over the baseline. These findings highlight the individual merits of MoMEB and UARB, as well as their complementary synergy when combined.

\begin{table}[t]
    \centering
    \caption{The performance of spectral-spatial expert mechanism on three benchmark datasets (reported in OA (\%)). The best results are highlighted in \textbf{Bold}. “w/o SRE/SSE” represents the absence of the routed spatial expert or shared spectral expert, while “w/ SRE \& SSE” indicates that dynamic spectral-spatial experts are employed.}
    \resizebox{\linewidth}{!}{
    \begin{tabular}{c c c c}
    \toprule
       \textbf{MambaMoE} &  Pavia University & Houston & Whu-HanChuan \\
        \hline
        w/o SRE    &   93.17   &   90.56   &  91.44   \\
        w/o SSE    &   94.24   &   90.31   &  91.78   \\
        \rowcolor{gray!30}  {w/ SRE \& SSE}    &   \textbf{95.20}   &   \textbf{91.18}   &  \textbf{92.67}  \\
    \bottomrule
    \end{tabular}
    }
    \label{table:ablation_2}
\end{table}

\begin{table}[t]
\caption{Effects of number of top-k experts on the three benchmark datasets (reported in OA (\%)). The best results are highlighted in \textbf{Bold}.}
\centering
\resizebox{\linewidth}{!}{
\begin{tabular}{c|ccc} %
\toprule
Number & Pavia University & Houston    & Whu-HanChuan \\ \hline
k=1    &     94.86  & 89.67 & 91.60        \\
k=2    &    95.03   & 90.54 & 91.71           \\
\rowcolor{gray!30} k=3    &   \textbf{95.20} & \textbf{91.18} & \textbf{92.67}\\
k=4    &    94.97 & 90.79 & 92.28\\
\bottomrule
\end{tabular}
}
\label{table:topk}
\end{table}

\noindent\textbf{(2) Internal Design of DSSEM.} To further investigate the internal architecture of the DSSEM in the proposed MoMEB module, we conduct an ablation study by isolating the contributions of the SRE and SSE. This analysis aims to evaluate the individual and combined impact of these components on classification performance. As shown in Table~\ref{table:ablation_2}, the removal of either SRE or SSE results in a consistent decline in accuracy across all three HSI datasets, highlighting the importance of jointly leveraging spatial and spectral contextual modeling. For example, on the Pavia University dataset, the full DSSEM achieves an OA of 95.20\%. In contrast, employing only SRE or only SSE leads to OAs of 94.24\% and 93.17\%, corresponding to performance drops of 0.96\% and 2.03\%, respectively. These findings affirm that the complementary interaction between SRE and SSE is crucial for effective spectral-spatial representation learning, thereby validating the design rationale of the DSSEM.

\begin{table*}[!htb]
{ 
\centering
\caption{Quantitative classification results of the Pavia University dataset. Best results are shown in \textbf{bold}.}
\label{Pavia-OA}
\resizebox{1\textwidth}{!}{
\begin{tabular}{cccccccccccc}
    \toprule
Class       & GAHT        & CSIL          & MASSFormer & DSFormer     & SS-Mamba            & HyperMamba  & S2Mamba     & MambaLG     & MambaHSI            & MambaMoE            \\ \hline
1           & 90.05±3.83  & 71.00±4.99    & 93.69±1.77 & 92.53±4.43   & \textbf{96.03±2.05} & 78.35±6.03  & 74.12±5.73  & 69.39±4.98  & 87.02±1.95          & 92.80±1.90          \\
2           & 87.42±5.40  & 87.48±4.65    & 87.48±8.38 & 90.23±4.15   & 85.95±5.47          & 77.73±2.97  & 59.82±15.46 & 79.31±3.20  & \textbf{94.55±1.37} & 93.39±2.05          \\
3           & 86.02±5.19  & 81.45±5.04    & 90.77±4.21 & 92.50±3.55   & 95.54±2.53          & 72.57±10.50 & 56.90±22.26 & 80.23±5.91  & 89.32±3.70          & \textbf{93.64±4.87} \\
4           & 95.64±0.76  & 84.81±4.17    & 95.79±1.45 & 96.71±0.96   & 97.70±0.84          & 93.97±2.03  & 94.95±1.43  & 94.28±1.27  & 79.36±1.35          & \textbf{96.94±0.82} \\
5           & 99.60±0.52  & 98.96±1.01    & 99.62±0.51 & 99.89±0.31   & \textbf{100}        & 99.92±0.15  & 99.90±0.01  & 99.91±0.11  & 99.67±0.20          & \textbf{100}        \\
6           & 93.31±4.76  & 95.91±2.75    & 94.38±5.17 & 96.52±3.13   & 94.17±7.47          & 73.92±3.66  & 58.70±11.57 & 81.98±7.43  & 98.52±0.87          & \textbf{99.82±0.43} \\
7           & 98.18±1.20  & 87.81±7.69    & 98.43±1.79 & 99.05±0.71   & \textbf{99.55±0.73} & 92.81±3.65  & 90.32±4.22  & 95.91±1.18  & 87.73±0.90          & 98.53±0.68          \\
8           & 93.62±3.86  & 75.50±9.07    & 91.65±4.01 & 91.90±4.41   & 93.26±4.44          & 77.65±9.09  & 79.00±18.31 & 88.12±5.43  & 90.51±3.36          & \textbf{98.13±1.28} \\
9           & 95.34±2.48  & 91.28±6.14    & 98.26±0.91 & 99.12±0.83   & \textbf{99.67±0.39} & 99.54±0.22  & 97.50±1.01  & 98.70±1.67  & 93.31±0.45          & 98.53±1.44          \\ \hline
OA (\%)     & 90.46±2.66  & 84.85±2.79    & 91.32±3.96 & 92.81±2.20   & 91.58±2.35          & 79.89±1.76  & 68.98±5.99  & 81.53±1.68  & 92.08±0.70          & \textbf{95.20±1.31} \\
$\kappa$ (\%)  & 87.67±3.29  & 80.40±3.41    & 88.81±4.84 & 90.66±2.78   & 89.13±2.92          & 74.21±2.05  & 61.57±6.31  & 76.41±2.06  & 89.57±0.90          & \textbf{93.73±1.32} \\
AA (\%)     & 93.24±1.11  & 86.02±1.78    & 94.45±1.58 & 95.38±1.15   & 95.76±1.27          & 85.09±0.83  & 79.02±2.23  & 87.54±1.38  & 91.11±0.55          & \textbf{96.87±0.57} \\
\bottomrule
\end{tabular}
}}
\end{table*}

\begin{figure*}[h]
\centering
\includegraphics[width=\textwidth]{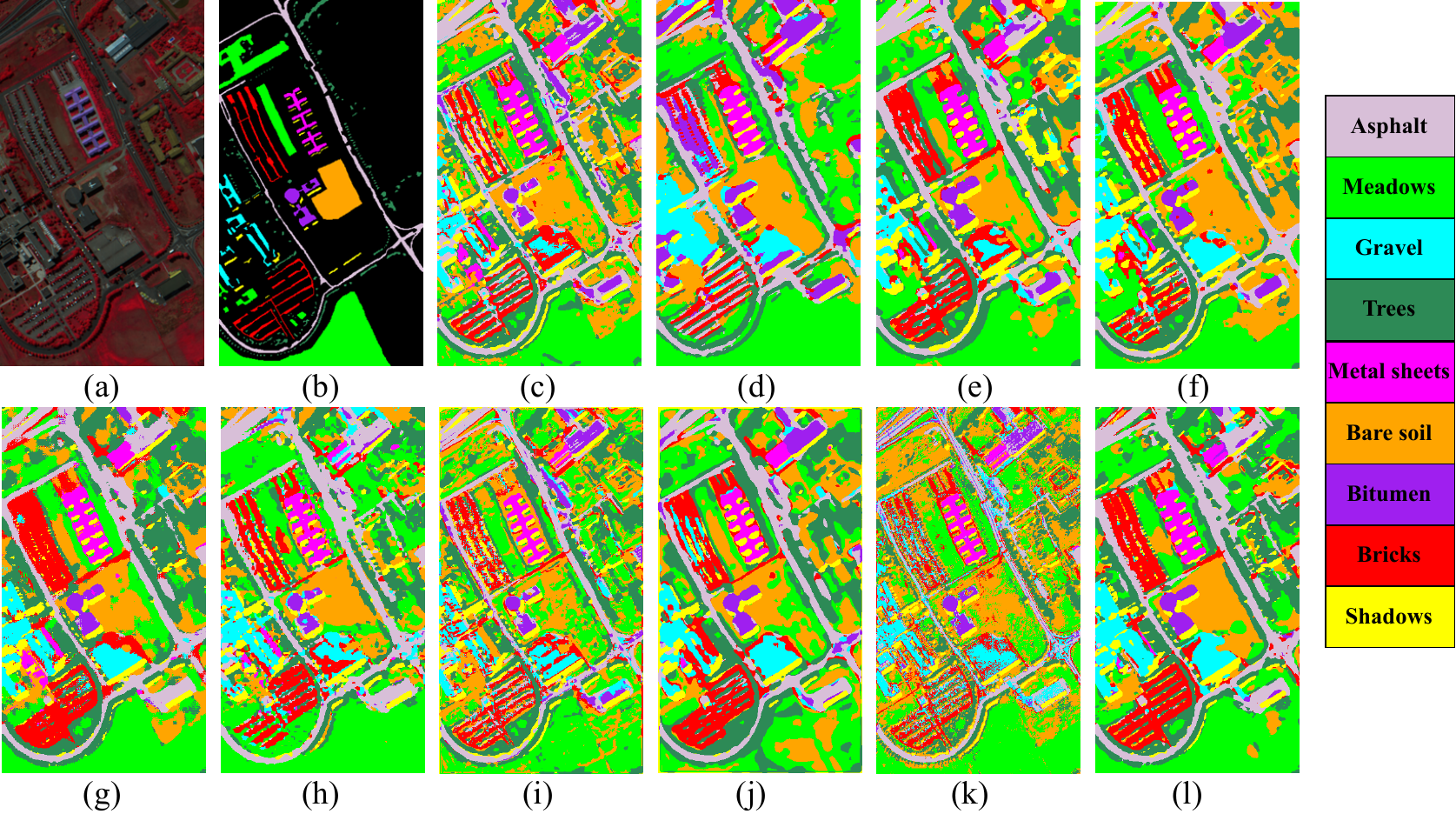}
\caption{Visualization of the classification maps on Pavia University dataset. (a) False-color image. (b) Ground truth. (c) GAHT. (d) CSIL. (e) MASSFormer. (f) DSFormer. (g) SS-Mamba. (h) HyperMamba. (i) S2Mamba. (j) MambaLG. (k) MambaHSI. (l) MambaMoE.}
\label{Pavia_map}
\end{figure*}

\noindent\textbf{(3) Number of Selected Experts in SRE.} After MoMEB and DSSEM, we further investigate the optimal configuration of the SRE, specifically focusing on the impact of varying the number of top-$k$ selected experts during inference. To this end, experiments were conducted on all three benchmark datasets to evaluate classification performance under different $k$ values. As reported in Table~\ref{table:topk}, the OA consistently improves as $k$ increases, with the best performance achieved at $k=3$ across all datasets. This performance gain is attributed to the enhanced spatial context modeling facilitated by incorporating a broader set of directional experts, which proves beneficial for handling the complexity and heterogeneity of ground objects in hyperspectral scenes. However, a noticeable decline in OA is observed when $k=4$, suggesting a diminishing return beyond a certain level of expert aggregation. This drop can be explained by the introduction of redundant or less relevant features, which potentially dilute the discriminative capacity of the learned representations. Moreover, when $k=4$, the model effectively reduces to a fully weighted combination of all experts, thereby losing the sparsity and dynamic routing characteristics fundamental to the MoE paradigm. These results underscore that selecting the top-3 experts during inference offers the most effective trade-off between spatial diversity and feature relevance, thereby yielding the highest classification accuracy.

\begin{table*}[t]
{ 
\centering
\caption{Quantitative classification results of the Houston dataset. Best results are shown in \textbf{bold}.}\label{Houston-OA}
\resizebox{1\textwidth}{!}{
\begin{tabular}{cccccccccccc}
    \toprule
Class     & GAHT                & CSIL        & MASSFormer   & DSFormer            & SS-Mamba            & HyperMamba          & S2Mamba             & MambaLG             & MambaHSI   & MambaMoE            \\ \hline
1         & 89.64±7.55          & 79.21±5.17  & 92.43±5.00   & 90.79±6.35          & 91.25±5.78          & \textbf{98.10±1.09} & 86.93±5.26          & 85.63±0.12          & 95.28±1.53 & 97.57±0.48          \\
2         & 91.46±5.55          & 63.42±11.16 & 93.96±4.91   & 94.55±5.38          & 93.81±5.35          & 93.37±8.91          & 56.59±15.70         & \textbf{97.30±1.79} & 92.81±0.88 & 96.06±1.38          \\
3         & 99.52±0.32          & 95.01±4.76  & \textbf{99.79±0.20}   & 96.55±1.66          & 99.30±0.24          & 99.06±1.04          & {99.72±0.23}          & 99.56±0.45          & 93.43±0.86 & {99.18±0.42} \\
4         & 90.81±2.74          & 52.59±8.56  & 92.94±3.99   & 94.75±2.52          & 94.24±2.56          & 92.02±2.60          & 92.96±3.35          & 93.98±0.40          & 96.95±0.95 & \textbf{97.30±0.30} \\
5         & 99.23±1.44          & 88.87±8.39  & 99.26±0.80   & 99.84±0.27          & 99.47±1.54          & 97.44±2.50          & 99.30±0.15          & 98.12±1.13          & 99.80±0.15 & \textbf{100}        \\
6         & 89.74±3.97          & 91.64±4.52  & 93.68±6.33   & 91.74±5.45          & 92.29±6.36          & 92.66±4.95          & 88.68±4.88          & 86.41±0.95          & 96.61±0.82 & \textbf{97.94±0.88} \\
7         & 90.05±3.78          & 57.78±6.51  & 87.06±7.71   & 87.29±4.57          & 89.66±5.32          & 77.86±9.59          & \textbf{90.54±3.54} & 76.56±18.36         & 83.85±2.13 & 84.33±1.29          \\
8         & 78.45±4.94          & 56.16±7.42  & 75.45±5.67   & 73.60±6.80          & \textbf{78.78±6.42} & 67.69±5.26          & 42.07±16.66         & 78.36±3.49          & 54.52±3.06 & 70.07±3.25          \\
9         & 87.30±2.75          & 53.96±9.29  & 86.69±3.90   & 82.45±3.24          & 86.55±3.92          & 76.55±6.80          & 80.41±5.78          & 70.87±4.32          & 78.46±1.86 & \textbf{87.49±2.40} \\
10        & 84.19±11.10         & 76.25±7.80  & 87.38±9.77   & 82.02±11.61         & 87.82±12.74         & 77.22±1.94          & 41.37±18.91         & 89.10±3.61          & \textbf{94.83±2.10} & {92.15±2.96} \\
11        & 83.76±6.43          & 74.37±5.02  & 82.11±7.35   & 81.30±4.56          & 83.98±7.39          & 74.20±2.07          & 68.34±11.31         & \textbf{87.73±3.12}          & 77.32±1.62 & {86.11±1.59} \\
12        & 81.89±5.96          & 65.18±6.23  & 83.82±6.02   & {83.99±8.83} & 75.80±10.90         & 72.93±8.88          & 48.14±19.35         & 83.84±4.68          & 78.89±3.93 & \textbf{85.73±4.84}          \\
13        & \textbf{96.70±2.20} & 79.42±6.82  & 95.04±1.46   & 96.21±1.65          & 95.62±3.70          & 88.10±3.41          & 94.12±1.22          & 60.17±5.87          & 91.87±4.58 & 94.54±2.39          \\
14        & 99.90±0.29          & 98.37±3.19  & 99.90±0.19   & 99.64±0.38          & 99.98±0.07          & 99.59±0.37          & 99.98±0.01          & 99.17±0.57          & 99.95±0.15 & \textbf{100}        \\
15        & 99.83±0.36          & 90.17±7.09  & \textbf{100} & \textbf{100}        & 99.88±0.22          & 98.59±1.32          & \textbf{100}        & 99.44±0.71          & 95.89±0.76 & \textbf{100}        \\ \hline
OA (\%)    & 89.43±1.16          & 70.94±1.83  & 89.83±1.59   & 88.80±1.72          & 89.83±2.22          & 85.04±0.80          & 75.28±2.92          & 86.89±2.10          & 86.96±0.52 & \textbf{91.18±0.61} \\
$\kappa$  (\%) & 88.58±1.25          & 68.66±1.96  & 89.01±1.72   & 87.90±1.86          & 89.01±2.38          & 83.86±0.87          & 73.33±3.15          & 85.83±2.26          & 85.91±0.56 & \textbf{92.56±0.54} \\
AA (\%)    & 90.83±0.87          & 74.89±1.64  & 91.30±1.50   & 90.32±1.53          & 91.23±1.68          & 87.03±0.97          & 79.28±2.22          & 87.08±1.71          & 88.70±0.38 & \textbf{92.24±0.83}\\        
\bottomrule
\end{tabular}
}}
\end{table*}

\begin{figure*}[h]
\centering
\includegraphics[width=\textwidth]{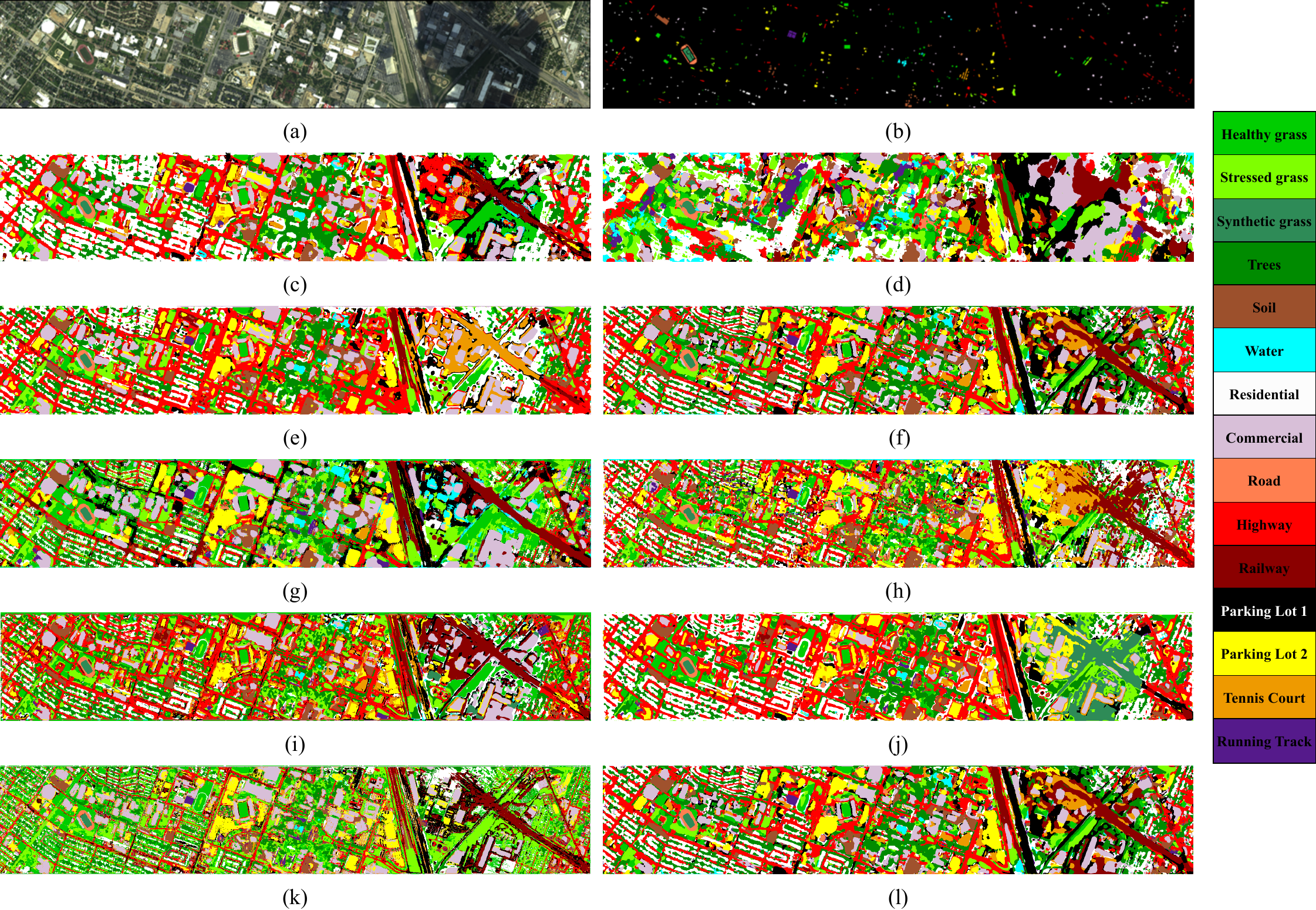}
\caption{Visualization of the classification maps on Houston dataset. (a) False-color image. (b) Ground truth. (c) GAHT. (d) CSIL. (e) MASSFormer. (f) DSFormer. (g) SS-Mamba. (h) HyperMamba. (i) S2Mamba. (j) MambaLG. (k) MambaHSI. (l) MambaMoE.
}
\label{Houston_map}
\end{figure*}

\subsection{Performance Comparison and Analysis}
In this subsection, we conduct a comprehensive comparison between the proposed MambaMoE and existing state-of-the-art (SOTA) HSI classification methods from both quantitative and qualitative perspectives. The compared methods fall into two main types: (1)  Transformer-based methods, including GAHT \cite{GAHT}, CSIL \cite{CSIL}, MASSFormer \cite{MASSFormer}, and DSFormer \cite{DSFormer}; and (2) Mamba-based architectures, including SS-Mamba \cite{huang-SSMamba}, MambaHSI \cite{MambaHSI}, HyperMamba \cite{hypermamba}, S2Mamba \cite{s2mamba}, and MambaLG \cite{mambaLG}.

\begin{table*}[!htb]
{ 
\centering
\caption{Quantitative classification results of the Whu-Hanchuan dataset. Best results are shown in \textbf{bold}.}\label{Hanchuan-OA}
\resizebox{1\textwidth}{!}{
\begin{tabular}{cccccccccccc}
    \toprule
Class     & GAHT       & CSIL                & MASSFormer          & DSFormer            & SS-Mamba    & HyperMamba & S2Mamba             & MambaLG    & MambaHSI            & MambaMoE            \\ \hline
1         & 86.95±3.58 & 84.56±3.39          & 89.98±3.19          & 90.38±3.52          & 86.00±7.62  & 85.21±5.95 & \textbf{96.05±4.16} & 72.16±6.66 & 96.04±0.58          & 95.20±1.57          \\
2         & 81.29±5.54 & 78.12±3.79          & 88.11±4.91          & 86.97±6.76          & 68.30±6.88  & 82.75±3.16 & 77.65±3.09          & 61.13±6.90 & 80.63±2.06          & \textbf{91.60±1.74} \\
3         & 80.61±8.70 & 90.72±5.37          & 90.87±5.04          & 89.83±5.12          & 79.03±9.94  & 84.37±3.52 & 77.72±8.44          & 86.08±4.95 & \textbf{97.69±0.36} & 89.54±10.56         \\
4         & 96.91±1.20 & 96.67±2.22          & \textbf{98.92±0.85} & 97.86±1.32          & 96.56±2.20  & 93.69±3.53 & 97.23±0.86          & 97.49±0.65 & 97.48±0.58          & 91.92±4.10          \\
5         & 97.85±1.26 & \textbf{99.98±0.05} & 98.52±2.32          & 98.26±1.53          & 92.88±7.25  & 97.58±2.28 & 99.69±0.21          & 93.04±0.64 & 98.66±0.52          & 98.72±0.55          \\
6         & 72.05±4.04 & 88.65±3.19          & 83.51±4.09          & \textbf{84.76±3.65} & 65.29±10.69 & 56.61±6.74 & 38.38±7.84          & 58.79±5.65 & 82.04±5.97          & 60.35±7.13          \\
7         & 87.55±7.02 & 87.87±4.09          & 91.00±4.40          & 92.19±3.94          & 89.62±6.16  & 92.30±2.10 & 95.10±1.08          & 87.55±4.25 & \textbf{96.70±0.97} & 81.84±11.65         \\
8         & 69.58±7.82 & 85.32±4.79          & 80.16±4.08          & 79.05±4.55          & 68.28±8.89  & 69.68±4.50 & 45.73±12.90         & 58.52±7.24 & 71.27±2.27          & \textbf{85.80±0.81} \\
9         & 80.58±5.22 & 78.62±4.56          & 83.51±5.22          & 89.74±3.92          & 71.88±8.46  & 75.74±7.67 & 29.82±12.24         & 58.71±8.31 & \textbf{92.33±1.55} & 89.38±6.40          \\
10        & 97.39±1.65 & 87.98±5.09          & 97.33±2.41          & \textbf{98.38±1.18} & 94.06±7.36  & 94.06±4.06 & 96.05±0.54          & 91.73±3.71 & 90.78±1.41          & 92.85±1.04          \\
11        & 90.56±5.54 & 76.54±5.78          & 92.17±9.12          & 94.84±1.57          & 76.54±24.24 & 91.09±4.21 & 92.89±2.89          & 88.33±5.41 & \textbf{96.79±1.26} & 95.63±1.66          \\
12        & 86.04±7.56 & \textbf{99.67±0.41} & 95.27±2.82          & 92.09±3.56          & 80.65±10.13 & 91.26±2.37 & 62.88±5.77          & 69.30±8.50 & 87.97±0.88          & 85.44±3.43          \\
13        & 77.79±5.89 & 81.94±4.11          & 79.00±6.55          & 82.49±5.01          & 65.82±10.77 & 75.93±4.70 & 57.79±3.78          & 50.16±6.65 & \textbf{89.36±0.41} & 75.22±3.21          \\
14        & 77.11±7.76 & 74.62±5.84          & 81.43±5.97          & 86.42±4.94          & 75.37±10.70 & 83.41±5.21 & 64.22±6.72          & 73.96±5.48 & 87.51±1.05          & \textbf{92.94±2.36} \\
15        & 96.65±3.16 & 94.24±2.33          & 96.48±2.12          & 98.37±2.12          & 94.18±5.36  & 92.74±2.90 & 98.86±0.62          & 91.62±2.79 & 93.88±0.44          & \textbf{99.94±0.04} \\
16        & 96.18±4.31 & 82.15±2.30          & 97.30±3.97          & 97.86±1.83          & 94.80±6.49  & 92.37±0.71 & 92.27±5.76          & 98.60±1.20 & 95.08±0.51          & \textbf{99.05±0.20} \\ \hline
OA (\%)    & 87.12±1.69 & 82.89±1.45          & 90.75±1.98          & 91.68±1.28          & 83.14±4.48  & 87.58±0.70 & 79.83±2.39          & 79.45±1.26 & 91.27±0.25          & \textbf{92.67±1.22} \\
$\kappa$  (\%) & 85.05±1.89 & 80.29±1.65          & 89.23±2.27          & 90.31±1.47          & 80.44±5.16  & 85.55±0.79 & 76.66±2.66          & 76.23±1.40 & 89.84±0.30          & \textbf{91.68±1.16} \\
AA (\%)    & 85.94±1.24 & 86.73±1.22          & 90.22±1.61          & \textbf{91.22±0.84} & 81.21±4.47  & 85.21±0.18 & 75.86±1.79          & 77.23±1.30 & 90.89±0.65          & 89.09±1.68\\        
\bottomrule
\end{tabular}
}}
\end{table*}

\begin{figure*}[h]
\centering
\includegraphics[width=\textwidth]{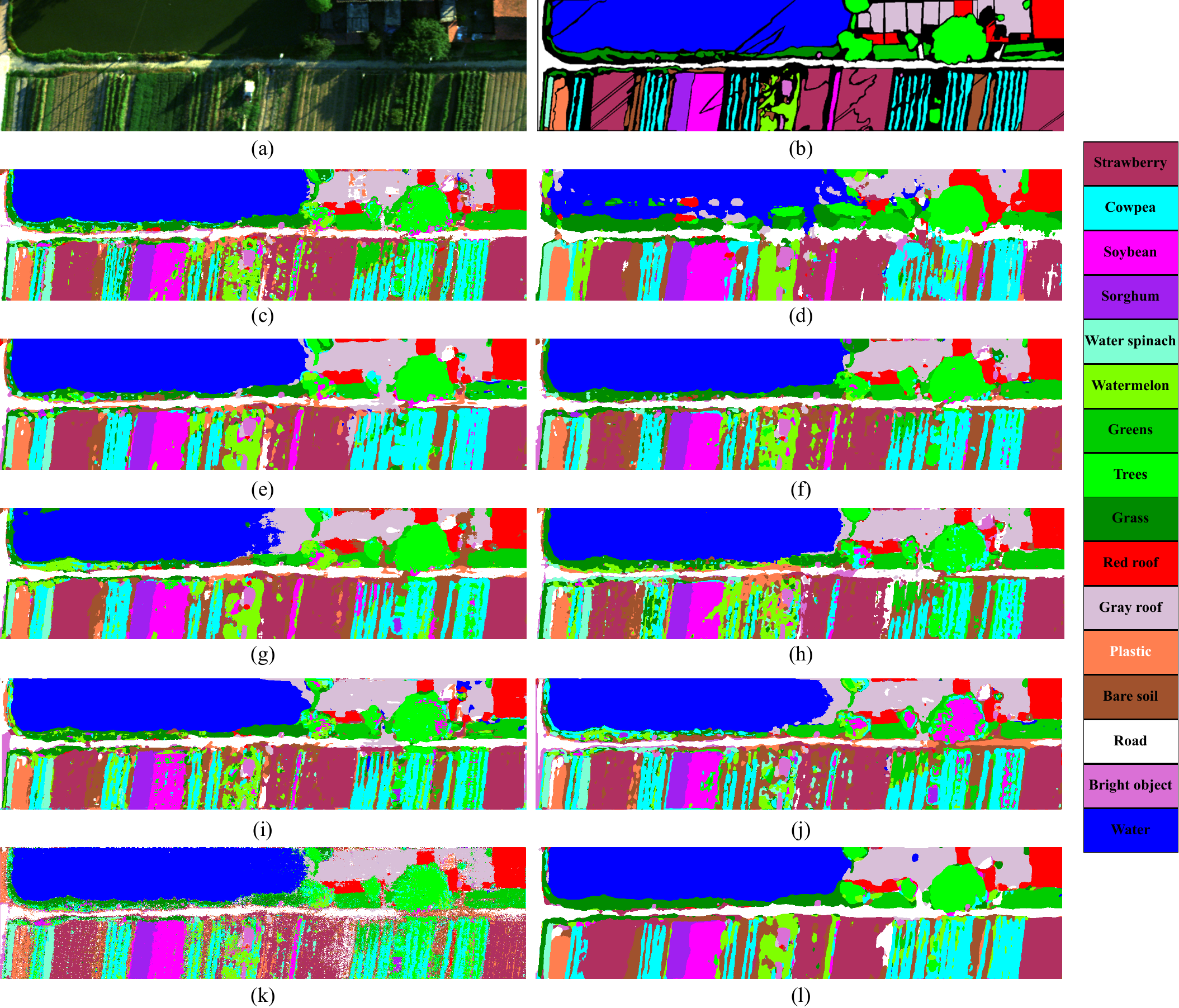}
\caption{Visualization of the classification maps on Whu-HanChuan dataset. (a) False-color image. (b) Ground truth. (c) GAHT. (d) CSIL. (e) MASSFormer. (f) DSFormer. (g) SS-Mamba. (h) HyperMamba. (i) S2Mamba. (j) MambaLG. (k) MambaHSI. (l) MambaMoE.}
\label{Hanchuan_map}
\end{figure*}

\noindent\textbf{(1) Quantitative Results.} The quantitative results are presented in Table~\ref{Pavia-OA} to Table~\ref{Hanchuan-OA}, highlighting the comparative performance of various state-of-the-art methods. Since MambaMoE integrates image-level feature extraction with dynamic spectral-spatial context modeling through adaptive directional feature extraction and sparse expert activation. Moreover, the UGCL strategy enhances model learning on challenging regions by enforcing stage-wise supervision, thereby mitigating the limitations of Mamba's inherently sequential modeling architecture. As a result, MambaMoE consistently outperforms all competing methods, including most Mamba-based approaches that rely on patch-level classification, which typically generate representations with limited receptive fields and struggle to capture the holistic semantics of remote sensing scenes. Even compared with MambaHSI, which also utilizes image-level input and achieves an OA of 92.08\% on the Pavia University dataset.

\noindent\textbf{(2) Qualitative Results.} We further present the classification maps in Fig.~\ref{Pavia_map} to Fig.~\ref{Hanchuan_map} to visually evaluate the performance of various methods. As illustrated, the proposed MambaMoE delivers superior classification results, characterized by smooth and coherent land-cover structures, well-defined boundaries, and minimal salt-and-pepper noise, closely aligning with the ground truth. Notably, the areas highlighted by the black boxes reveal that methods such as SSFCN, SpectralFormer, HyperMamba, and MambaLG suffer from considerable salt-and-pepper artifacts. Additionally, models like SACNet, FcontNet, DSFormer, and MambaHSI exhibit issues such as misclassification and blurred boundaries. In contrast, MambaMoE accurately delineates the contours of the “Bitumen” class and achieves precise classification of categories like “Bricks” and “Asphalt”, preserving both spatial consistency and semantic clarity. 

\noindent\textbf{(3) Robustness with fewer training samples.} To rigorously evaluate the stability of the proposed MambaMoE architecture, we conducted additional experiments under more challenging conditions using fewer training samples. Specifically, in addition to the standard experimental setup, we created training subsets containing only 5 and 10 samples per class for each of the three benchmark datasets. Subsequently, we compared the OA of MambaMoE with several strong baselines: DSFormer, S2Mamba and MambaHSI, which were selected based on their competitive performance. The experimental results reveal that MambaMoE consistently outperforms all competitors, even with severely limited supervision. Notably, the drop in performance remains minimal, demonstrating that MambaMoE retains effectiveness despite reduced training data. These findings underscore the robustness, generalization capability, and practicality of MambaMoE in real-world scenarios where labeled hyperspectral data is often scarce.

\begin{figure}[t]
    \centering
     \includegraphics[width=\linewidth]{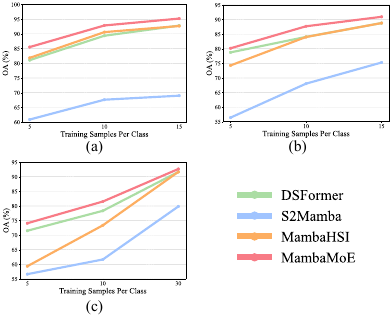}
     \caption{Comparison of classification performance with fewer training samples: (a) Pavia University dataset. (b) Houston dataset. (c) Whu-HanChuan dataset.}
     \label{fig:sample}
\end{figure}

\noindent\textbf{(4) Computational Cost Analysis.} We conducted a comprehensive evaluation of the proposed MambaMoE against several Transformer- and Mamba-based baselines, focusing on four critical metrics: model parameters, floating-point operations (FLOPs), training time, and inference time. The comparison results are summarized in Table~\ref{computational_efficiency}. 
It can be seen that, although MambaMoE is not lightweight in terms of model parameters, it still achieves the second-lowest FLOPs and inference time among all compared methods, demonstrating a well-balanced trade-off between classification accuracy and computational efficiency. For instance, on the Pavia University dataset, MambaMoE achieves an OA of 95.20\% (see Table \ref{Pavia-OA}) with an inference time of only 0.21 seconds. This advantage primarily stems from the sparse activation mechanism of the MoE architecture, where only a subset of expert modules is engaged during each forward pass, substantially reducing redundant computation. Furthermore, the use of image-level inputs allows for parallel inference across a large number of samples, further boosting efficiency. In summary, MambaMoE offers high performance with relatively low computational overhead, making it particularly suitable for resource-constrained or real-time hyperspectral remote sensing applications.

\begin{table*}[!ht]
\centering
\caption{Comparison of computational efficiency among different methods on the Pavia University dataset. The number of model parameters and FLOPs are measured with an input size of $103 \times 13 \times 13$. The best and second-best results are highlighted in \textbf{bold} and underlined, respectively.}\label{computational_efficiency}
\begin{tabular}{lccccccc}
\toprule
\rowcolor{gray!20}  Metrics & GAHT & MASSFormer & DSFormer & S2Mamba & MambaLG & MambaHSI & MambaMoE \\
\midrule
Parameters (M) & 0.93 & 0.30 & 0.59 & 0.14 & \underline{0.13} & \textbf{0.03} & 0.28 \\
FLOPs (M) & 156.62 & 51.02 & 60.25 & 24.79 & 30.96 & \textbf{5.35} & \underline{12.13} \\
Training time (s)  & 40.23 & \textbf{19.29} & 57.93 & 42.06 & 955.01 & 95.86 & \underline{34.97} \\
Inference time (s) & 44.11 & 6.88  & 62.17 & 3.67  & 4.32   & \textbf{0.13}   & \underline{0.21} \\
\bottomrule
\end{tabular}
\end{table*}

\subsection{Visualization} 

\begin{figure}[t]
    \centering
     \includegraphics[width=1\linewidth]{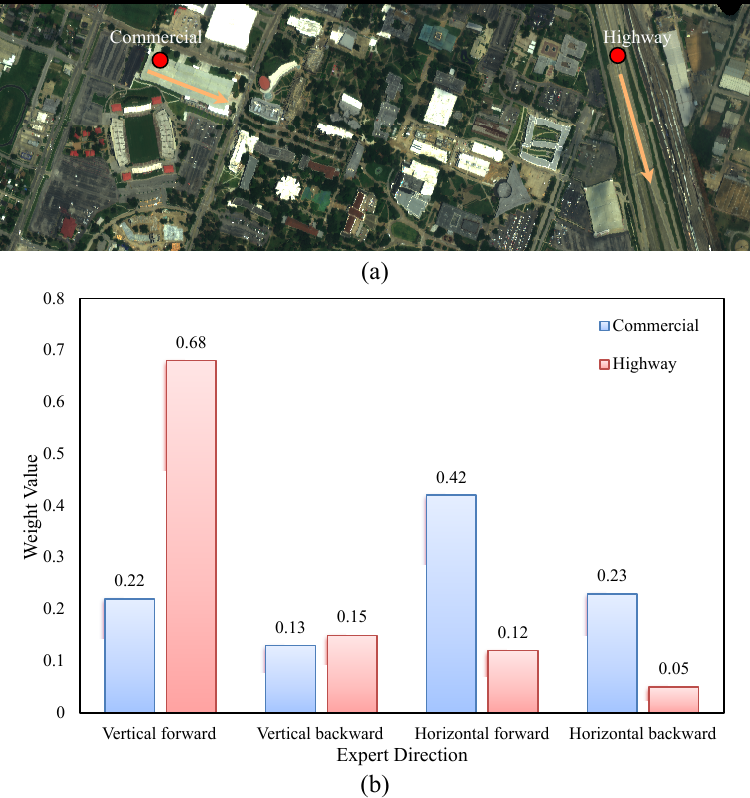}
     \caption{Visualization of directional expert weights for different land-cover types on the Houston dataset.}
     \label{expert_w}
\end{figure}

\textbf{(1) Validation of Direction Modeling.} \label{vis} To further validate the necessity of heterogeneous directional modeling for different land-cover types, we selected two representative categories in the Houston dataset with distinct orientation characteristics: commercial buildings and highways (see Fig.~\ref{expert_w}(a)), for detailed analysis. We examined the spatial expert weights in the proposed MambaMoE to investigate how directional patterns influence expert activation. Fig.~\ref{expert_w}(b) shows the expert weight distributions across four directional scanning paths: vertical forward, vertical backward, horizontal forward, and horizontal backward. It can be observed that the highway class exhibits a clearly vertical structure, with vertically oriented experts contributing a combined weight of 0.83, while horizontally oriented experts account for only 0.17. In contrast, the commercial building class displays a strong horizontal layout, with horizontal experts contributing 0.65 and vertical experts the remaining 0.35. For land-cover types with more complex or ambiguous spatial structures, effective contextual modeling relies on the collaborative contribution of all four directional experts, highlighting the model’s ability to adaptively integrate diverse directional cues.

\begin{figure}[t]
    \centering
     \includegraphics[width=0.93\linewidth]{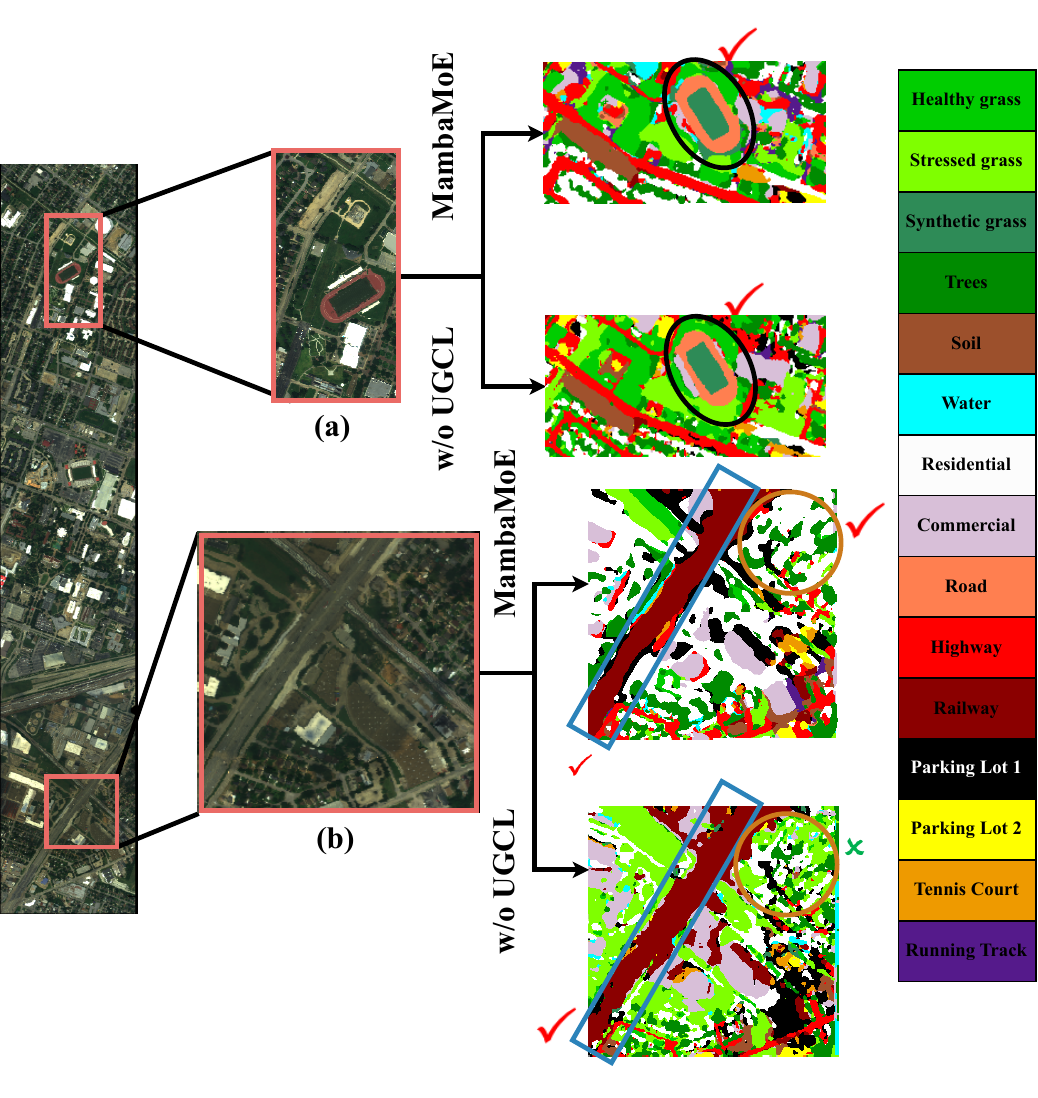}
     \caption{Comparative visualization of MambaMoE and its variant without UGCL on the Houston dataset. }
     \label{fig:UGCL}
\end{figure}

\textbf{(2) Specific Analysis of UGCL Component.} To further validate the necessity of the UGCL component within the proposed MambaMoE framework, we conducted a series of visualization analyses using the Houston dataset. As depicted in Fig.~\ref{fig:UGCL}, two representative regions with differing classification complexities were selected for comparative evaluation.

Region (a) represents a playground area comprising primarily roads and synthetic grass, characterized by clear boundaries and a homogeneous composition. In this relatively simple context, the MoMEB module alone proves sufficient to dynamically extract effective spectral-spatial features, resulting in accurate classification. This highlights MoMEB’s strong discriminative ability for land-cover types with distinct spatial regularity and minimal ambiguity.

In contrast, Region (b) encompasses both the relatively easy-to-distinguish railway class and the more challenging residential class, which exhibits complex, heterogeneous patterns. The classification results show that, similar to Region (a), MoMEB can accurately classify well-defined structures such as railways. However, in the intricate residential area, the absence of the UARB module often results in residential samples being misclassified as stressed grass. When UARB is incorporated, the classification accuracy within this complex region improves substantially. This observation underscores the effectiveness of UARB in enhancing the model’s capacity to differentiate complex and ambiguous land-cover types.

In summary, while MoMEB excels in handling straightforward categories with clear spatial features, its modeling capacity is limited in the presence of intricate or semantically diverse land-cover structures. To overcome this limitation, the UGCL strategy integrates the UARB module into the decoder, guiding MoMEB outputs through dynamic label refinement based on predictive uncertainty. This targeted supervision significantly boosts the model’s ability to capture nuanced semantic patterns in challenging HSI scenes.

\section{Conclusion}\label{Section5}
In this paper, we propose MambaMoE, a novel spectral-spatial mixture-of-experts framework for HSI classification. MambaMoE adaptively extracts joint spectral-spatial features that are tailored to the diverse characteristics of land covers in challenging HSI scenes. Specifically, the proposed MoMEB consists of spatially routed experts and spectrally shared experts, enabling dynamic learning of directional spectral-spatial features through a sparsely MoE mechanism. Additionally, we introduce an uncertainty-guided corrective learning (UGCL) strategy at the decoder stage. This strategy explicitly encourages the model to focus more on complex regions, alleviating classification confusion caused by the inherent limitations of directional modeling in Mamba. Comprehensive evaluations on three HSI benchmark datasets demonstrate that MambaMoE achieves superior performance in both classification accuracy and computational efficiency, validating the effectiveness of the proposed modules and training strategy.


\bibliographystyle{elsarticle-num-names}
\bibliography{MambaMoE}


\end{sloppypar}
\end{document}